\documentclass[a4paper, 10 pt, conference]{ieeeconf}

\IEEEoverridecommandlockouts                              

\usepackage{xcolor}

\definecolor{MyDarkBlue}{rgb}{0,0.08,1}
\definecolor{MyDarkGreen}{rgb}{0.02,0.6,0.02}
\definecolor{MyDarkRed}{rgb}{0.8,0.02,0.02}
\definecolor{MyDarkOrange}{rgb}{0.40,0.2,0.02}
\definecolor{MyPurple}{rgb}{111,0,255}
\definecolor{MyRed}{rgb}{1.0,0.0,0.0}
\definecolor{MyGold}{rgb}{0.75,0.6,0.12}
\definecolor{MyDarkgray}{rgb}{0.66, 0.66, 0.66}

\usepackage{graphics} 
\usepackage{epsfig} 
\usepackage{mathptmx} 
\usepackage{times} 
\usepackage{amsmath} 
\usepackage{amssymb}  
\usepackage{color}
\usepackage{gensymb}
\usepackage{siunitx}
\usepackage{comment}

\newcommand{\myparagraph}[1]{\vspace{0.1in}\noindent\textbf{#1}}

\title{\LARGE \bf GelSight Baby Fin Ray: A Compact, Compliant, Flexible Finger with High-Resolution Tactile Sensing}

\author{
    \authorblockN{Sandra Q. Liu, Yuxiang Ma, and Edward H. Adelson}
        \authorblockA{Massachusetts Institute of Technology\\
    {\tt\small sqliu@mit.edu, yxma@csail.mit.edu, adelson@csail.mit.edu}} 
}

\usepackage{multirow}
\usepackage{comment}

\begin{document}

\maketitle
\thispagestyle{empty}
\pagestyle{empty}

\begin{abstract}

The synthesis of tactile sensing with compliance is essential to many fields, from agricultural usages like fruit picking, to sustainability practices such as sorting recycling, to the creation of safe home-care robots for the elderly to age with dignity. From tactile sensing, we can discern material properties, recognize textures, and determine softness, while with compliance, we are able to securely and safely interact with the objects and the environment around us. These two abilities can culminate into a useful soft robotic gripper, such as the original GelSight Fin Ray \cite{og_finray}, which is able to grasp a large variety of different objects and also perform a simple household manipulation task: wine glass reorientation. Although the original GelSight Fin Ray solves the problem of interfacing a generally rigid, high-resolution sensor with a soft, compliant structure, we can improve the robustness of the sensor and implement techniques that make such camera-based tactile sensors applicable to a wider variety of soft robot designs. We first integrate flexible mirrors and incorporate the rigid electronic components into the base of the gripper, which greatly improves the compliance of the Fin Ray structure. Then, we synthesize a flexible and high-elongation silicone adhesive-based fluorescent paint, which can provide good quality 2D tactile localization results for our sensor. Finally, we incorporate all of these techniques into a new design: the Baby Fin Ray, which we use to dig through clutter, and perform successful classification of nuts in their shells.

\end{abstract}

\section{INTRODUCTION}
As both the fields of soft robotics and high-resolution tactile sensing continue to progress, we look towards progressing their intersection and advancing the field of soft manipulation. Soft robots are useful for a myriad of tasks ranging from human-robot interaction for care of the elderly, biomedical devices and prosthetics, and agricultural usages. However, soft robots and manipulators can become more useful when they are paired with intricate and high-resolution sensing capabilities \cite{wang2018toward}. Despite the ability of soft robotic mechanisms to comply to different objects and thus provide a more secure grasp, they are unable to perform more interesting manipulation tasks. This lack of ability can be partially solved with tactile sensing. 

\begin{figure}[ht!]
	\centering
	\includegraphics[width=1.0 \linewidth]{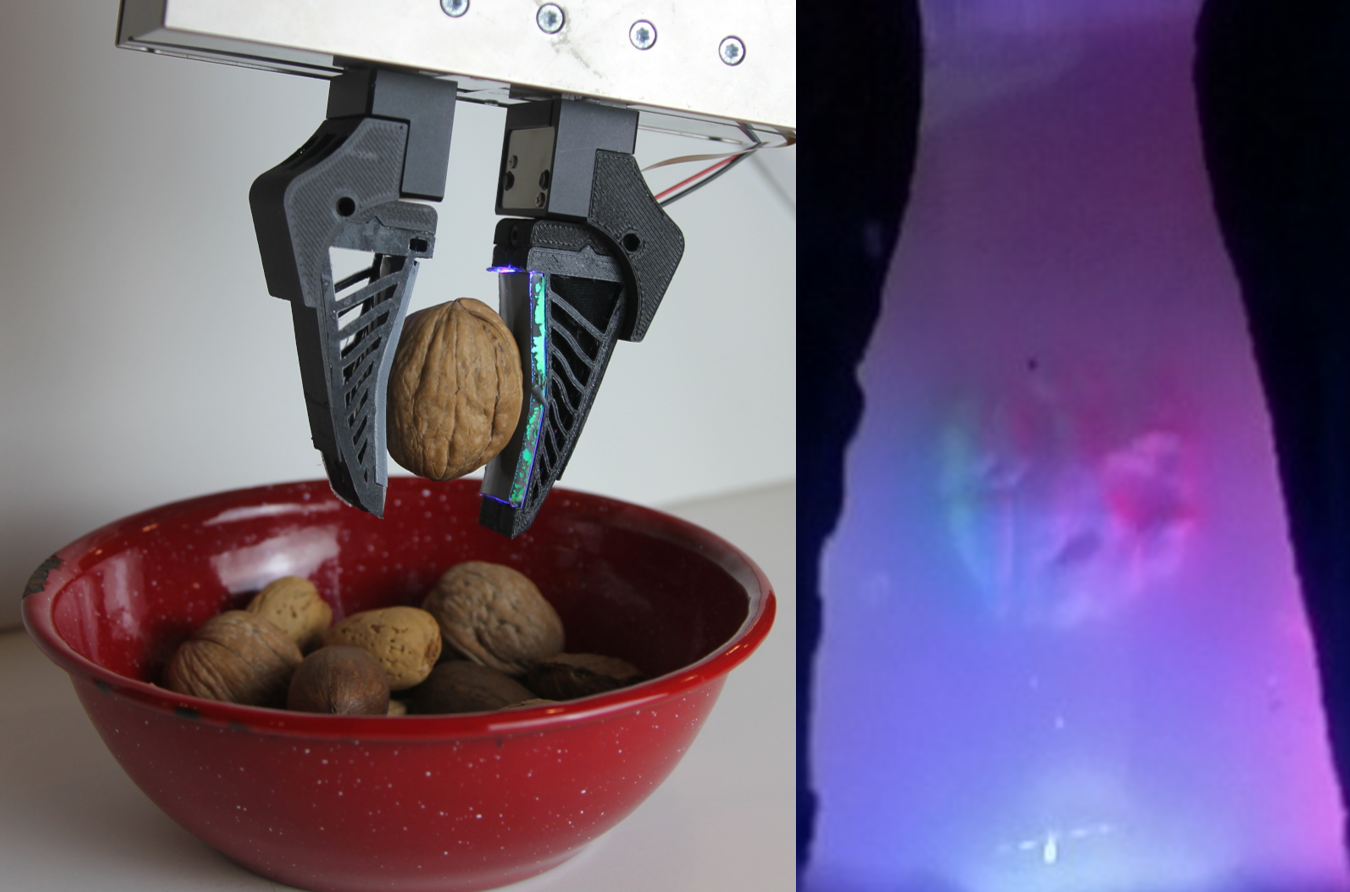}
    \vspace{-20pt}
	\caption{The Baby Fin Ray grabbing a walnut (left) and the corresponding cropped raw image of the mirror that displays the tri-colored tactile sensing region and the indentation of the walnut shell (right).}
	\vspace{-8pt}
	\label{fig:teaser}
\end{figure}

Tactile sensing in humans allows us to perform multiple tasks that many soft robots cannot do, such as discerning shapes, materials, and textures from a single touch \cite{mechanoreceptors}. Although proprioception and simple force sensing are useful, these types of sensors cannot distinguish between different intricate textures, which can be useful in helping the elderly with tasks such as digging through their bag to find their keys or looking through a bag of nuts to find a walnut. 

We want the safety and compliance given to us by soft robotic gripper designs combined with the high-resolution tactile sensing that camera-based sensors can provide us. However, the integration of these normally rigid sensors into soft robotics and materials is difficult. The development of the original GelSight Fin Ray \cite{og_finray} has helped improve this field. Even so, there are existing issues with its design such as its lack of mechanical repeatability and a slight loss of compliance due to  sensor integration. To further solve this integration problem, we present the following contributions:

\begin{itemize}
    \item A novel design of a soft, compliant and robust GelSight sensor in the Baby Fin Ray (Fig. \ref{fig:teaser});
    \item Synthesis and analysis of fluorescent pigment that has potential use in other soft robotic integration of camera-based tactile sensors;
    \item Successful performance of nut classification (i.e. identifying textures) using the Baby Fin Ray.
\end{itemize}

\section{RELATED WORK}
\subsection{Fin Ray Grippers}
Fin Ray fingers are a family of soft grippers that take advantage of the Fin Ray Effect. They were originally inspired by the deformation of fish fins, which bend against the direction of the applied force \cite{pfaff2011application}. The Fin Ray has a simple structure consisting of two long fin bones and several horizontal ribs, which are 3D printable and easy to modify \cite{crooks2016fin, tawk2019fully}. Over the years, the Fin Ray structure has been successfully applied to many robotic tasks using its adaptive geometry \cite{crooks2016fin, manoonpong2022fin}. 

As a result, researchers have tried to predict the behavior of Fin Ray fingers and improve their performance on grasping tasks. To guide the structure design and optimize their compliance, Armanini et al. proposed a mathematical model based on the discrete Cosserat approach to predict the in-plane and out-of-plane stiffness \cite{armanini2021discrete}, while Shan et al. developed a pseudo-rigid-model to evaluate the grasping quality \cite{shan2020modeling}. Studies have also used the Finite Element Method (FEM) to simulate and optimize different Fin Ray designs \cite{tawk2019fully, chen2019bio}. Based on FEM simulations, Deng et al. established a hand-object database to determine the optimal design parameters of a Fin Ray gripper \cite{deng2021learning}. 

Besides structure optimization, there have also been attempts to add additional components to augment the mechanical performance of Fin Ray fingers. One common issue of Fin Ray fingers is that the fingers experience out-of-plane motion, which can be mitigated by inserting small rigid rods at two ends of the ribs \cite{xu2021compliant}. Crooks et al. also attached side supports and fingernails to enhance out-of-plane rigidity and enable the fingers to grip small objects \cite{crooks2016fin}. Friction pads or electroadhesive pads can also increase the gripping force and improve the grasping payload \cite{shin2021universal, chen2019bio}. 

Until recently, most research has been focused on the mechanical behavior of Fin Ray fingers; there is very limited work done on integrating Fin Rays with sensors. Yang et al. embedded pressure force sensors inside of Fin Ray fingers \cite{yang20213d}. Xu et al. realized intrinsic force sensing with a camera and a neural network to estimate forces using displacements \cite{xu2021compliant}. Although both of these works incorporate force feedback, there remains a need for high resolution tactile information to perceive and manipulate various objects.

\subsection{Camera-based Tactile Sensing}
Camera-based tactile sensing is a technique that takes high resolution images of a contact pad and works seamlessly with computer vision algorithms and image neural networks. Compared with conventional strain/force sensors, camera-based sensors are able to provide rich geometrical information about the contact areas \cite{yuan2017gelsight}. 

Although camera-based tactile sensors have mainly been embedded in rigid fingers or fingertips, there are already some promising attempts to combine camera-based tactile sensing with soft grippers. She et al. embedded cameras in an exoskeleton-covered soft finger and enabled both proprioceptive sensing and contact tactile sensing \cite{she2020exoskeleton}. Amini et al. integrated an Optitrack system and camera-based tactile sensing with a soft end effector in order to map surface roughness for sanding processes \cite{amini2020uncertainty}. In Faris et al.'s work, proprioception and exteroception of soft fingers are enabled by tracking internal marker patterns with a neuromorphic event-based camera \cite{faris2022proprioception}. 

The original GelSight Fin Ray embodied both the passive adaptability of Fin Ray fingers and the high-resolution tactile sensing of camera-based sensors \cite{og_finray}. It was able to separate proprioceptive changes from the tactile interaction between the silicone tactile sensing pad and objects, but the incorporation of the acrylic piece and the camera are unfavorable for the compliance of the finger. Moreover, the robustness of tactile sensing was not guaranteed under large deformations due to limited sensing area and relative movement between the LEDs and silicone pad.

\section{METHODS}

\begin{figure}[ht]
	\centering
	\includegraphics[width=0.9\linewidth]{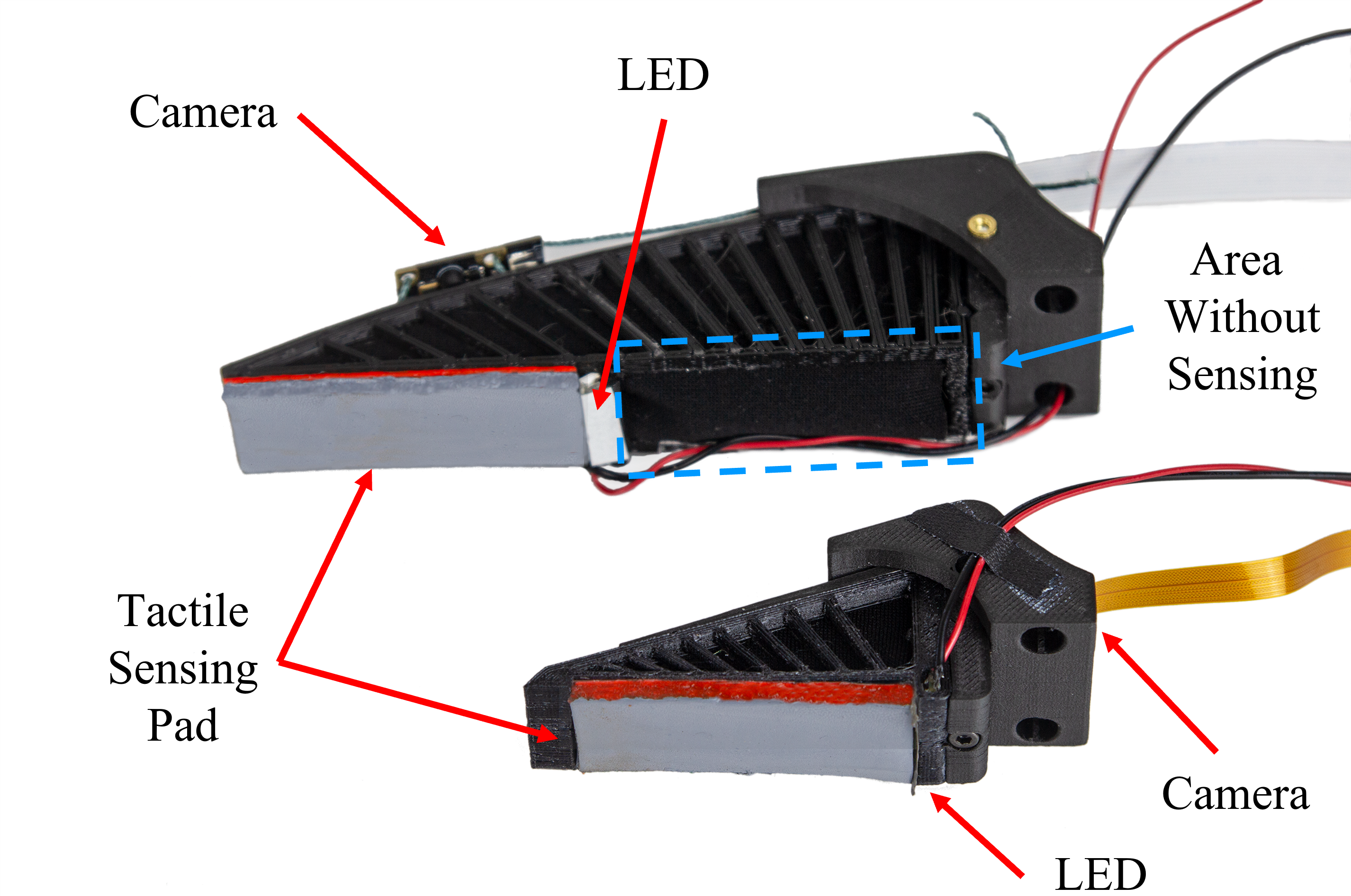}
    \vspace{-10pt}
	\caption{A side-by-side comparison of the original GelSight Fin Ray and the Baby Fin Ray.}
	\vspace{-5pt}
	\label{fig:comparison}
\end{figure}

\begin{figure}[ht]
	\centering
	\includegraphics[width= 0.99\linewidth]{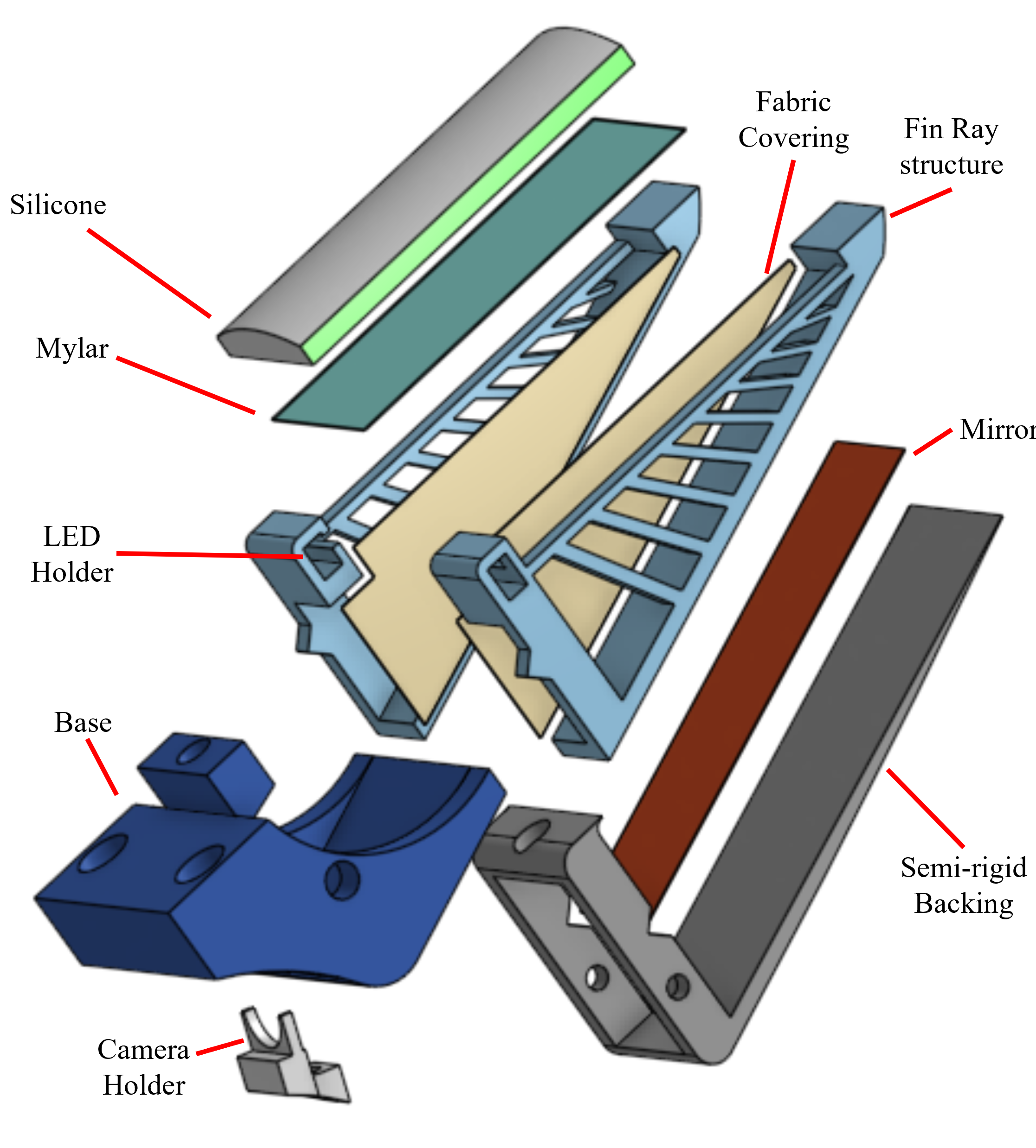}
    \vspace{-20pt}
	\caption{An exploded view of the Baby Fin Ray assembly. Not pictured are the camera, LEDs, and the diffuser.}
	\vspace{-10pt}
	\label{fig:explosion}
\end{figure}

The improvements to the original GelSight Fin Ray encompass mainly hardware updates and result in the development of the Baby Fin Ray. A side-by-side comparison of the two fingers are also shown in Fig. \ref{fig:comparison}, and the exploded version of the Baby Fin Ray is shown in Fig. \ref{fig:explosion}.

These advancements were made with the goal to improve and advance the design elements of the sensorized Fin Ray itself, while also showing more of its potential capabilities for tasks that require rich tactile sensing. Another important aspect is the improvement of techniques that could be utilized in other types of soft, compliant camera-based tactile sensors.

\subsection{Hardware}
Much of the hardware design for the Baby Fin Ray is inspired by the GelSight Fin Ray paper \cite{og_finray}. Like the original design, we create a hollow inner structure inside of the finger to allow unobstructed viewing of the tactile sensing surface. We utilize a 3D printed 1.75 mm TPU 95A material (Sainsmart) for the Fin Ray struts with a more rigid backing (Onyx material, Markforged) to prevent unwanted twisting motions in the finger that would be induced by hollowing out the structure. 

For the silicone gel pad, we use a mixture of 1 to 10 to 3 parts of XP-565 Parts A and B (Silicones Inc.) and a plasticizer (LC1550 Phenyl Trimethicone, Lotioncrafter), respectively. The XP-565 silicone is used because of its translucency. However, because the silicone itself is rigid, the plasticizer is added to soften the material and make it more suitable as a tactile sensor. 

Our mold is 3D printed out of Onyx material and a thin piece of 6 mil (0.15 mm) mylar is adhered to the bottom curved portion of the mold with cyanoacrylate glue. The mold has a 2.5 mm by 18 mm rectangular cross sectional area with a 25 mm radius curved profile along one of the longer sides. To ensure levelness of the sensing pad, a flat piece of translucent acrylic sheet is placed on top of the mold, leaving an area in which the silicone could be poured. The whole ensemble is secured with rubber bands and tilted at a slight angle to allow air pockets to rise to the top of the pouring pocket. Once the bubbles escape out of the mold, the mold assembly is placed inside of an oven and cured for 5 hours at 52$^{\circ}$C (125$^{\circ}$F). 

Afterwards, we synthesize our sensing membrane paint, which covers the curved surface of the silicone gel pad. This paint, like the one for the original GelSight Fin Ray, is composed of 1 part silicone ink catalyst to 10 parts gray silicone ink base to 2.5 parts 4 $\mu$m aluminum cornflakes to 30 parts NOVOCS Gloss (Raw Materials Inc., Schlenk, Smooth-on Inc). This is brushed only on top of the silicone gel pad, so that fluorescent paint can be applied to the sides of the silicone gel and so that the camera can see through the gel volume to the sensing surface. 

However, the Baby Fin Ray differs from its predecessor in many aspects. Specifically, there were a few problems with the design integration of the GelSight sensor into the Fin Ray in the original design. Despite its ability to comply to different object shapes while performing high-resolution tactile sensing, which made it able to do simple tactile manipulation tasks, the hardware was not very robust. The sensing region was limited to only half of the sensor despite the use of a large, wide angle fisheye lens camera, and the finger was unable to comply as much as a non-sensorized Fin Ray would be able to. These issues were addressed by our updated hardware design. The main problems we chose to fix included the following: 

\begin{itemize}
    \item Design for Usability
    \item Camera Placement
    \item Illumination \& Paint
\end{itemize}

Solving these problems also gives us the future potential to incorporate some or all of these methods in the development of other soft sensors or robots integrated with high-resolution camera-based sensors. 

\myparagraph{Design for Usability}
Despite the ability of the original GelSight Fin Ray to perform touch-based manipulation tasks, such as wine glass reorientation, it is not able to robustly perform many different types of tasks. Firstly, the LEDs are attached to the deformable acrylic piece that houses the silicone gel pad, meaning that if the Fin Ray grasps a large object or grasps it incorrectly, the LEDs can potentially be dislodged over time. Another issue is that the silicone gel pad is extended to the tip of the sensor, meaning that if we wanted to use the Fin Ray to grab an item out of clutter, the gel pad could potentially be dislodged. Finally, the long length of the Fin Ray (110 mm) makes it impossible to be viewed by a single camera.

As such, we halve the length of the Fin Ray design so that it is coincidentally around the size of a human finger. This length reduction serves two purposes: (1) the entire front of the sensor can be visible by a single camera; (2) one single illumination source is needed for the entire sensing region. 

We also place the blue LEDs (Chanzon 2835 SMD LEDs) at the base of the finger, protectively encased inside the flexible structure. Doing so further allows us to place a piece of VViViD Air-Tint Dark Air-Release Vinyl Wrap Film over the LEDs as a diffuser, which helps to prevent bright spots in the sensing region that were prevalent in the original design. 

The silicone gel pad is recessed into the Fin Ray structure so that there is less of a non-smooth interface between the tip of the Fin Ray and the silicone gel pad. We also sharpen the Fin Ray tip to somewhat emulate a fingernail. These changes serve to make the Baby Fin Ray more useful for a different variety of tasks, such as digging through a cluttered, obstructed environment to grab an object. 

\myparagraph{Camera placement}
One of the most prevalent issues with the GelSight Fin Ray is the large obtrusive camera on the backbone. Although this camera provides high-resolution wide-angle imaging, it also prevents the back of the Fin Ray from flexing. This obstruction makes it difficult for the overall structure to then comply to objects. 

A way to resolve this issue is to use a smaller camera that then corresponds with a smaller Fin Ray structure. However, this solution does not completely solve the problem of introducing unnecessary rigidity in the back of the structure, so we turn to the usage of flexible mirrors with inspiration from GelSlim and the GelSight Wedge \cite{gelslim, wedge}.

\begin{figure}[ht]
	\centering
	\includegraphics[width=1.01 \linewidth]{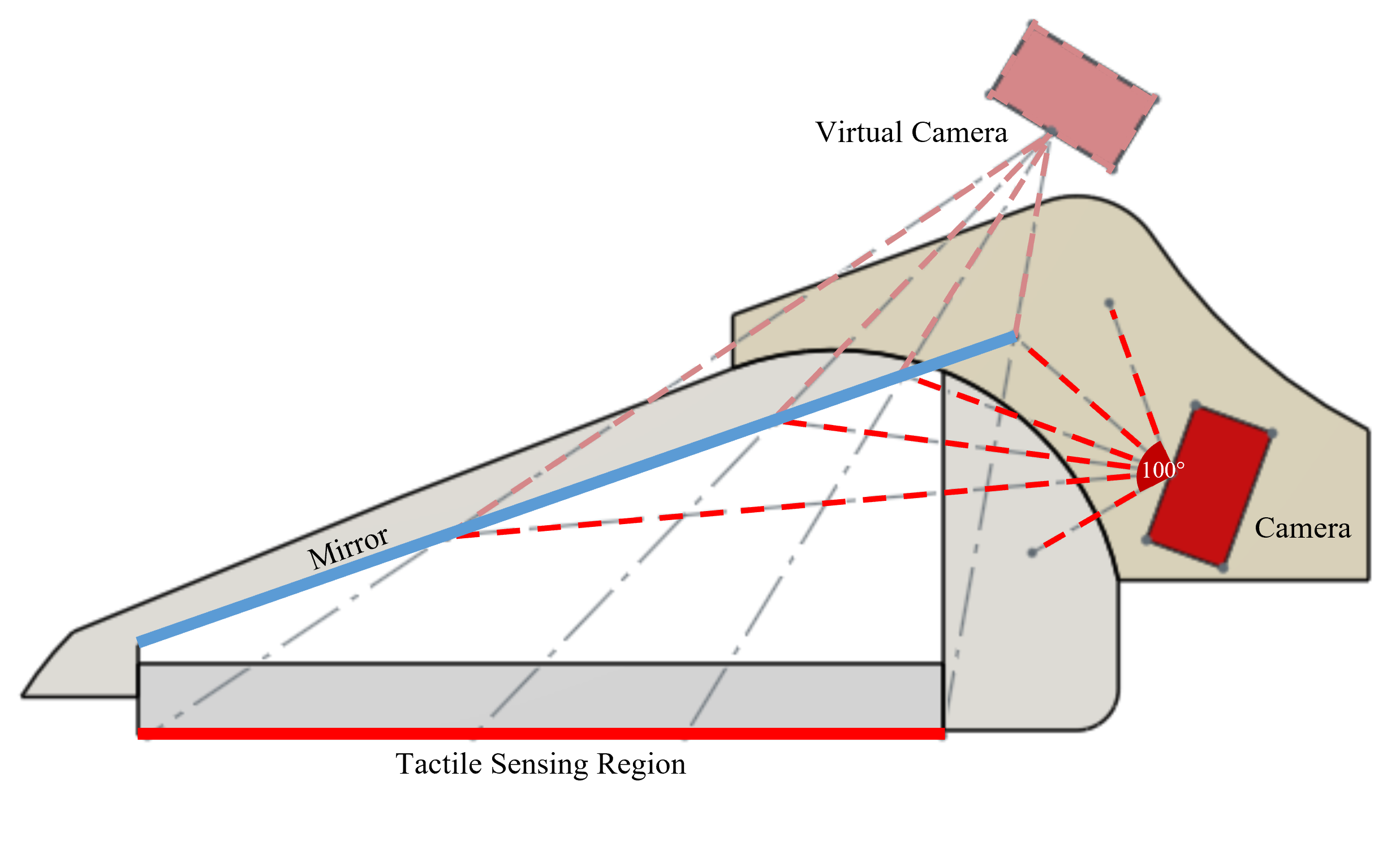}
    \vspace{-10pt}
	\caption{The 2D representation of a 100$^{\circ}$ field-of-view camera viewing the tactile sensing region through a mirror and its corresponding virtual camera. As the ray tracing shows, the camera is able to see the entire tactile sensing region while sitting at the base of the finger, where its rigidity will not impede with the deformable Fin Ray structure.}
	\vspace{-15pt}
	\label{fig:raytracing}
\end{figure}

In particular, we attach a 0.2 mm PET sheet to the inner back surface of the Fin Ray structure and place a camera at the base of the structure. The camera is angled in a way such that it can unobtrusively see the entire tactile sensing surface using the reflective surface of the PET sheet. Additionally, because the PET sheet is so thin, it can conform with the flexibility of the back structure, and we can see in Fig. \ref{fig:raytracing}, using CAD software, that we only need to utilize a 100$^{\circ}$ field-of-view camera to see the entire tactile sensing region. 

As such, the camera that we use for our design is a Raspberry Pi Zero Spy Camera with a 120$^{\circ}$ field of view. This camera is small enough to fit within the base, and it also provides a suitable viewing range for our tactile sensing region. Then, a yellow filter (Rosco E-Colour 765 Sunlight Yellow) is placed under the lens of the camera to help filter out the blue light, which could overpower the rest of the colors in the sensing region.

\myparagraph{Illumination \& Paint} Another issue we had to resolve was the slightly rigid acrylic piece in the original design, which serves to house the paint and provide a semi-rigid, deformable tactile sensing surface. The acrylic piece allows the acrylic fluorescent paint to deform with and illuminate the Fin Ray structure without the paint delaminating from silicone. This delamination occurs because the acrylic paint and silicone do not bond well together, necessitating the addition of an acrylic substrate. However, this addition increases the rigidity of the finger. 

We choose to completely eliminate the need for such an acrylic piece for this improved design. Not only does this improve upon the flexibility, it can also provide potential design strategies for non-LED illumination systems to other soft camera-based sensors. 

To be able to create a flexible, fluorescent paint that can bond to silicone, we use two different types of paint: a commercial fluorescent silicone pigment and a silicone paint of our own design. We choose to focus on adding as much pigment as possible to the fluorescent silicone mixtures so that we can have a more vivid tri-color sensing region. As a result, we do not optimize for the elongation, tensile strength, or other potentially relevant properties of the silicone paint. 

For the commercial fluorescent silicone pigment, we use Smooth On's Silc Pig Electric Green and Pink pigments. Following the instructions provided, we add only 3\% of the total silicone system weight to our mixture of XP-565 and NOVOCS Gloss (11 to 3), before we stir it, degas it, and paint it onto the long sides of the silicone gel pad in multiple layers. The NOVOCS Gloss is added to the mixture to help thin it out to make it easier to paint onto the silicone piece. The silicone gel pad is then placed into the oven to cure for 4 hours at 52$^{\circ}$C (125$^{\circ}$F). 

In an attempt to create a customizable flexible paint, which could work with multiple types of acrylic paint, we also synthesize a paint using a silicone adhesive base. In a small container, we measure out 0.5 grams of acrylic paint (Liquitex Basics Acrylic Paint Fluorescent Green and Red) and 1.5 grams of A564 medical adhesive silicone (Factor II, Inc). Because silicone adhesive cures faster with moisture and acrylic paint is water-based, we make sure that the silicone adhesive and paint are not in contact with one another as we add both of them to our container. 

Once both have been added, we vigorously stir until the mixture is mostly homogeneous and immediately add 1.0 gram of NOVOCS Gloss to thin out the mixture while constantly stirring. After the clumps of acrylic paint and silicone adhesive have dissolved into the NOVOCS Gloss, we slowly add 3.0 more grams of NOVOCS gloss in 1.0 gram increments, stirring after each addition. Doing so allows the paint to slowly thin out until it can be more easily applied to the sides of the silicone gel pad. To avoid large clumps in the paint, the mixture is strained over a 190 Micron filter (TCP Global) and then multiple layers of the paint are applied onto the silicone gel pad sides. The mixture is left to cure at room temperature for 30 minutes. We also note that this exact same procedure can be done with Silpoxy (Smooth On), and can potentially be performed with other silicone adhesives. However, we choose the A564 for its clearer coloring as opposed to other silicone adhesives that have more of a hazy quality, which we believe could limit the paint fluorescence. 

\subsection{Software}
The camera is connected to a Raspberry Pi and video is streamed using mjpg-streamer. For utilizing the data for tasks, we use a warp perspective function on the visible mirror in the image to convert it into a rectangular image with the correct ratio of the actual flexible mirror. This accounts for slight differences in manufacturing that might lead to changes in the ``negative" space around the flexible mirror and tactile sensing region. 

\section{ANALYSIS}
\subsection{Simulation}
To improve the compliance of the original GelSight Fin Ray and the robustness of hardware when doing manipulation, we make a more compact design by removing the acrylic piece and by changing the camera location. In order to prove that the new design has better compliance, we did a series of FEM simulations using ABAQUS software. We used continuum shell elements for thin sheets like the Mylar and flexible mirror, while we used eight-node linear brick elements for the main structures of the Fin Ray, e.g. silicone, semi-rigid backing, base, and the Fin Ray structure. Because of the large displacement, a non-linear analysis step was adopted to simulate the complex jamming behavior between the Fin Ray ribs.  

\begin{figure}[ht]
	\centering
	\includegraphics[width=1. \linewidth]{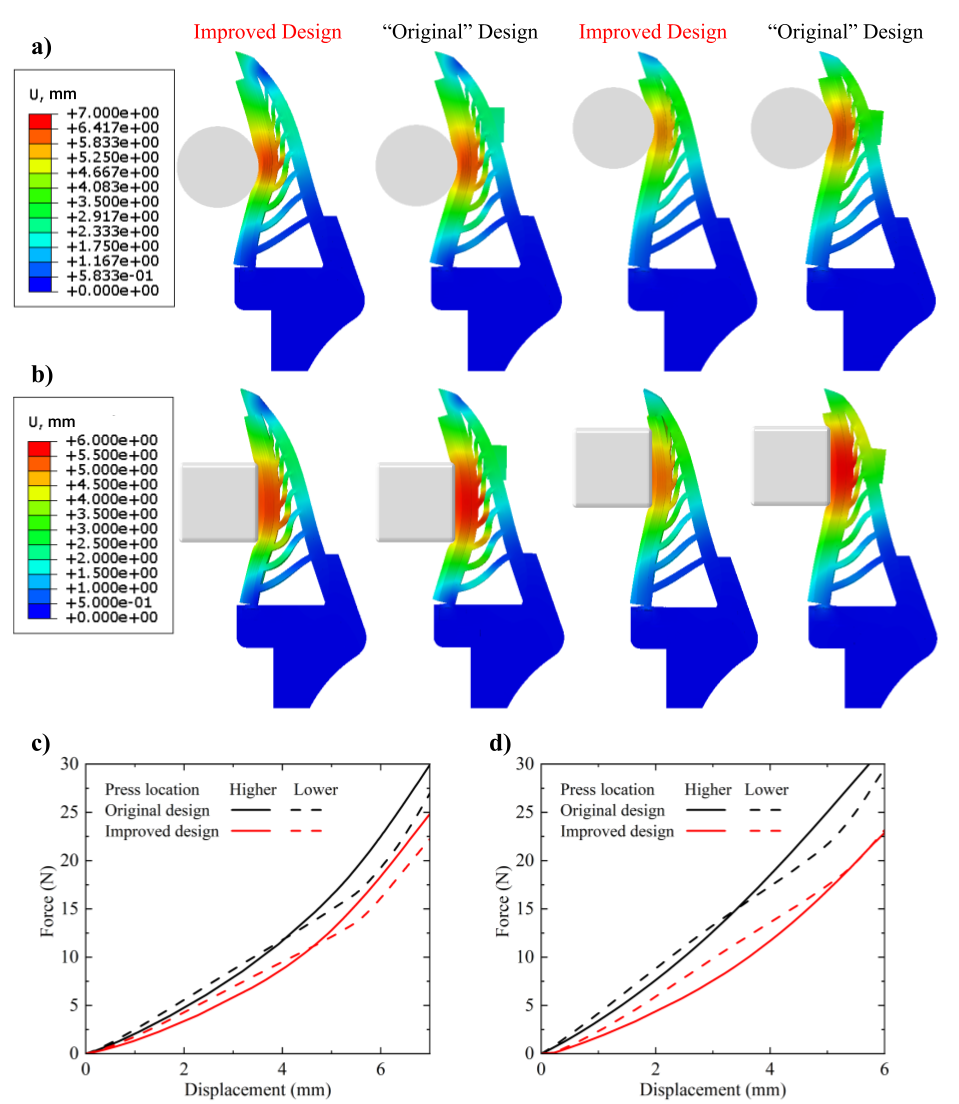}
    \vspace{-10pt}
	\caption{FEM simulation results of the improved design and the ``original" design, which is a smaller version of the previous GelSight Fin Ray for the sake of comparison. The indentations happen at the middle finger pad (lower location) and the fingertip a) Displacement fields of cylinder indentation; b) Displacement fields of cuboid indentation; c) Force-displacement curves of the improved and ``original" fingers with cylinder indenter; d) Force-displacement curves of the improved and ``original" fingers with cuboid indenter.}
	\label{fig:simulation}
	\vspace{-8pt}
\end{figure}

As shown in Fig. \ref{fig:simulation}, we make a comparison between the Baby Fin Ray and the ``original" Gelsight Fin Ray. The ``original" design is a smaller and comparable version of the previous original GelSight Fin Ray with the rigid acrylic piece inside the gel pad and the camera on the backbone. A cylinder indenter (shown in Fig. \ref{fig:simulation} a) and a cuboid indenter (shown in Fig. \ref{fig:simulation} b) are pressed into two fingers at two different locations, middle finger pad (lower location) and fingertip (higher location). Fig. \ref{fig:simulation} c and d show the Force-displacement curves of the two fingers pressed by the cylinder indenter and the cuboid indenter respectively. Depending on the shape of the indenter, indenter location, and depth, the Fin Ray fingers can display very different stiffness or compliance, but the improved design in general needs less pushing force to get the same amount of indentation depth. Overall, the Baby Fin Ray is 18-32\% more compliant than the original design. 

\subsection{Fluorescent Silicone Paint}
To analyze our synthesized fluorescent materials and compare them with one another and to the original acrylic paint on an acrylic piece, we performed tensile testing and created our own metric to analyze the illumination and 2D localization through reconstruction of each lighting scheme.

\myparagraph{Tensile Testing} 
Tensile testing was performed on the two types of synthesized silicone-based fluorescent paint. We chose not to perform this test on the dried acrylic paint since acrylic paint is brittle once dried. 

\begin{figure}[ht]
	\centering
	\includegraphics[width=0.5 \linewidth]{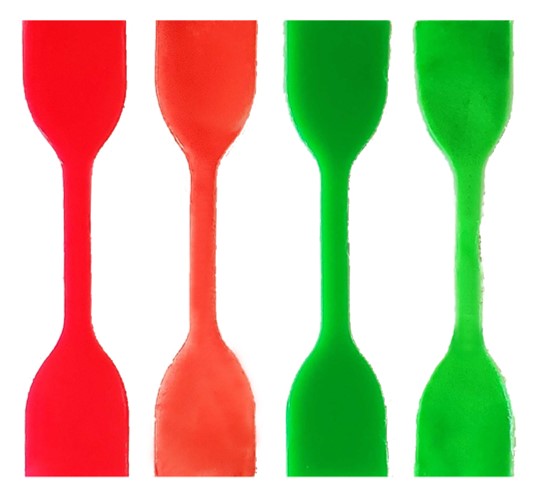}
    \vspace{-5pt}
	\caption{Fluorescent silicone paint dogbone samples used for tensile testing on the Instron meeting. From left to right, we have our pink Silc Pig, pink synthesized paint, green Silc Pig, and green synthesized paint samples.}
	\label{fig:bones}
\end{figure}

Test samples were prepared in a dogbone mold based on the ASTM D412 standard. The molds were printed with Onyx material on the Markforged printer. A laser cut 0.15 mm Mylar piece was placed on the bottom of the mold and an acrylic piece was clamped on the top of the mold to enforce an even thickness across the entire dogbone specimen. After preparing the samples mixtures, the mixtures were poured into the molds and left to cure per the paint preparation instructions. Some examples of our samples are shown in Fig. \ref{fig:bones}. 

After curing, all of the paint samples experienced some shrinkage. Although the dimensions of the dogbone profile of each piece was close to the others (sub-millimeter differences), the thicknesses varied a bit more (1 mm differences). As such, the dimensions were re-measured after curing and right before the dogbones were put into the tensile testing machine. Testing was performed on an Instron machine and samples were stretched until they broke. 

\myparagraph{Illumination/2D Localization} 
To test the integrity of our illumination schemes using different paints, we compared the reconstruction outputs (2D localization) of each paint illumination pattern against our ``base case": the acrylic paint on a deformable acrylic piece. 

We created three different tactile sensing samples and placed them on a Baby Fin Ray. Three different objects, a 5 mm by 10 mm rectangular block, a 4.75 mm diameter ball bearing, and a Lego block piece were then lightly pressed into the three different sensing regions. Corresponding reference images, images where no tactile imprint was given to the sensing region, were used to obtain a difference image that only showed the tactile imprint of our objects. This difference image was then used to obtain a reconstruction image, which we used as a 2D localization metric for comparing the position of the pressed object with the image reconstruction to determine the paint fidelity with respect to a GelSight sensor. 

This comparison was made by first manually segmenting the ball bearing and rectangular block. We then used a distance error metric for the center of the ball segmentation and utilized the Dice coefficient metric to compare the similarities of the rectangular segmentation image with the 2D localization results \cite{dice}. A spherical object and rectangular object were chosen to best represent the lighting illumination changes where both a smooth and a discrete object were pressed into the sensing region. In particular, we chose a spherical object to see if the lighting illumination was uniform enough for our reconstruction to determine the appropriate placement of the circle center. The Lego block piece was chosen to determine, in a qualitative way, how finely detailed the reconstruction image could be. However, for many manipulation tasks, it is unnecessary to have extremely high-resolution sensing that would allow us to see the tiny words on a Lego block piece, which are approximately 0.2 to 0.3 mm in width. 

\myparagraph{Paint Analysis Results}

\begin{table*}[!htb]
    \centering
    \caption{Fluorescent Paint Comparisons}
    \begin{tabular}{cccccc} \hline
    Name & Lego Pic. \& Reconstruction & Circle Seg. Error & Rect. Dice Score & Elongation @ Break & UTS \\ \hline
    Acrylic on Acrylic
    & \parbox[c]{10em}{\includegraphics[width=10em]{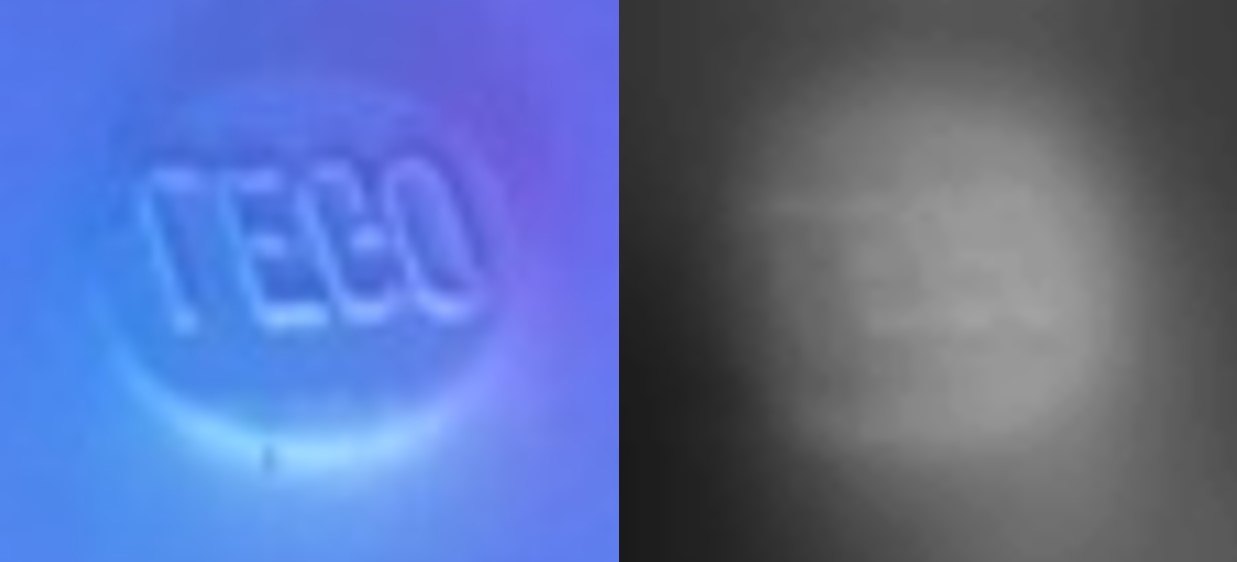}} &
    3.6 pixels   & 72.3\%   & N/A & N/A   \\
    Acrylic + Silicone Adhesive
    & \parbox[c]{10em}{\includegraphics[width=10em]{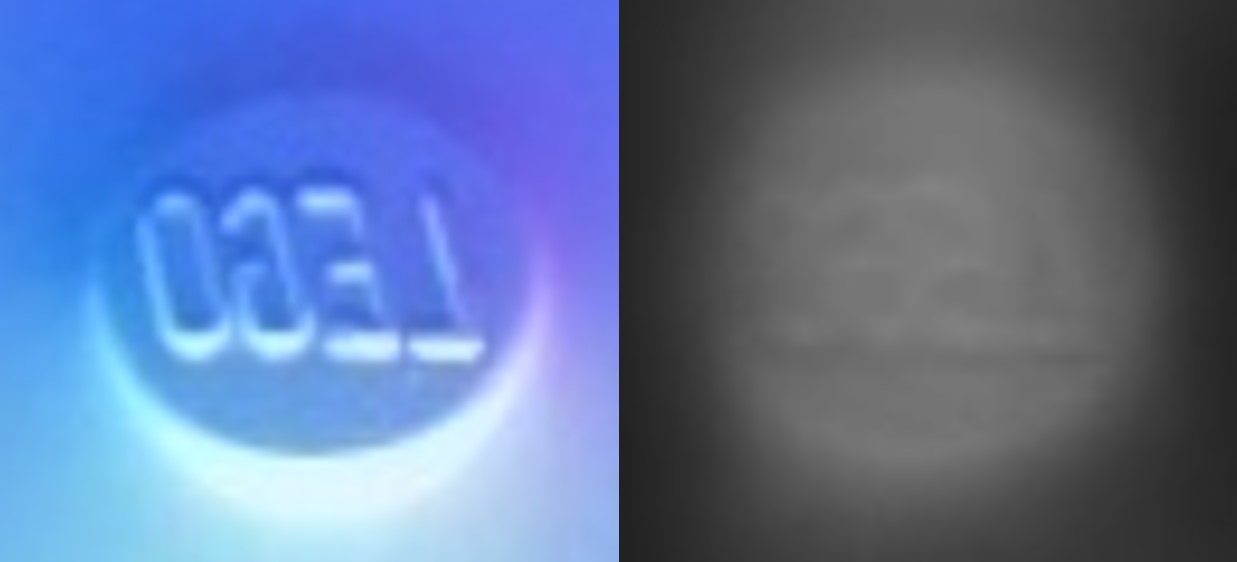}}  & 
    1.4 pixels   & 85.0\%   & 1,212\%  & 1.37 MPa  \\
    Silc Pig Pigment  & \parbox[c]{10em}{\includegraphics[width=10em]{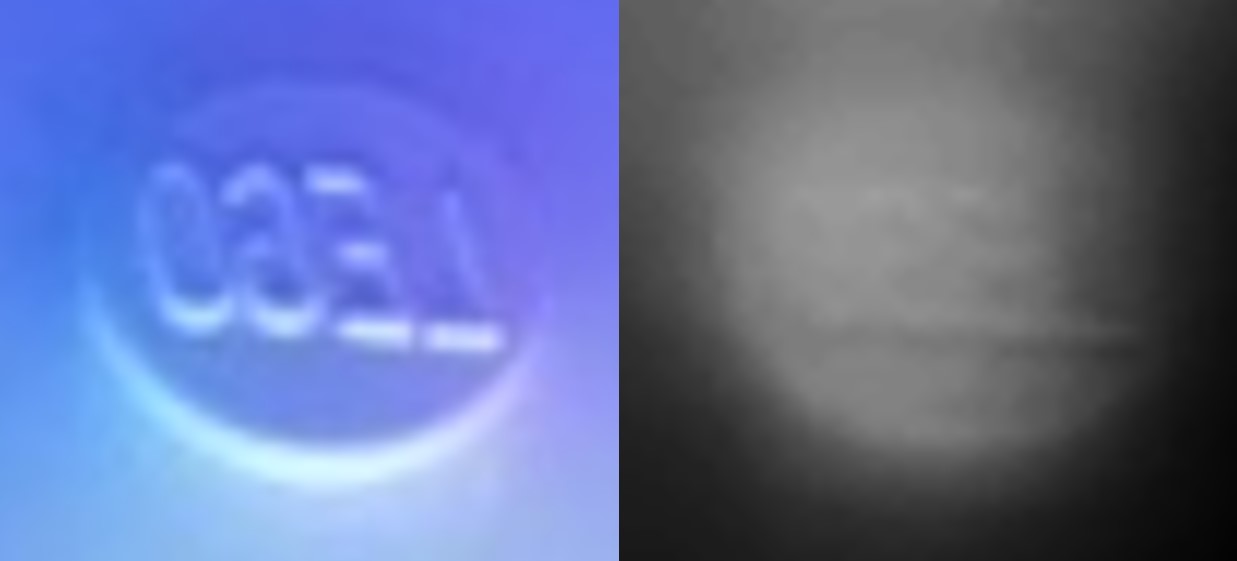}} & 3.6 pixels & 75.0\%   & 267\% & 0.23 MPa \\  \hline \label{tab:bleh} \end{tabular}
    \end{table*}

\begin{figure}[ht]
	\centering
	\includegraphics[width=1. \linewidth]{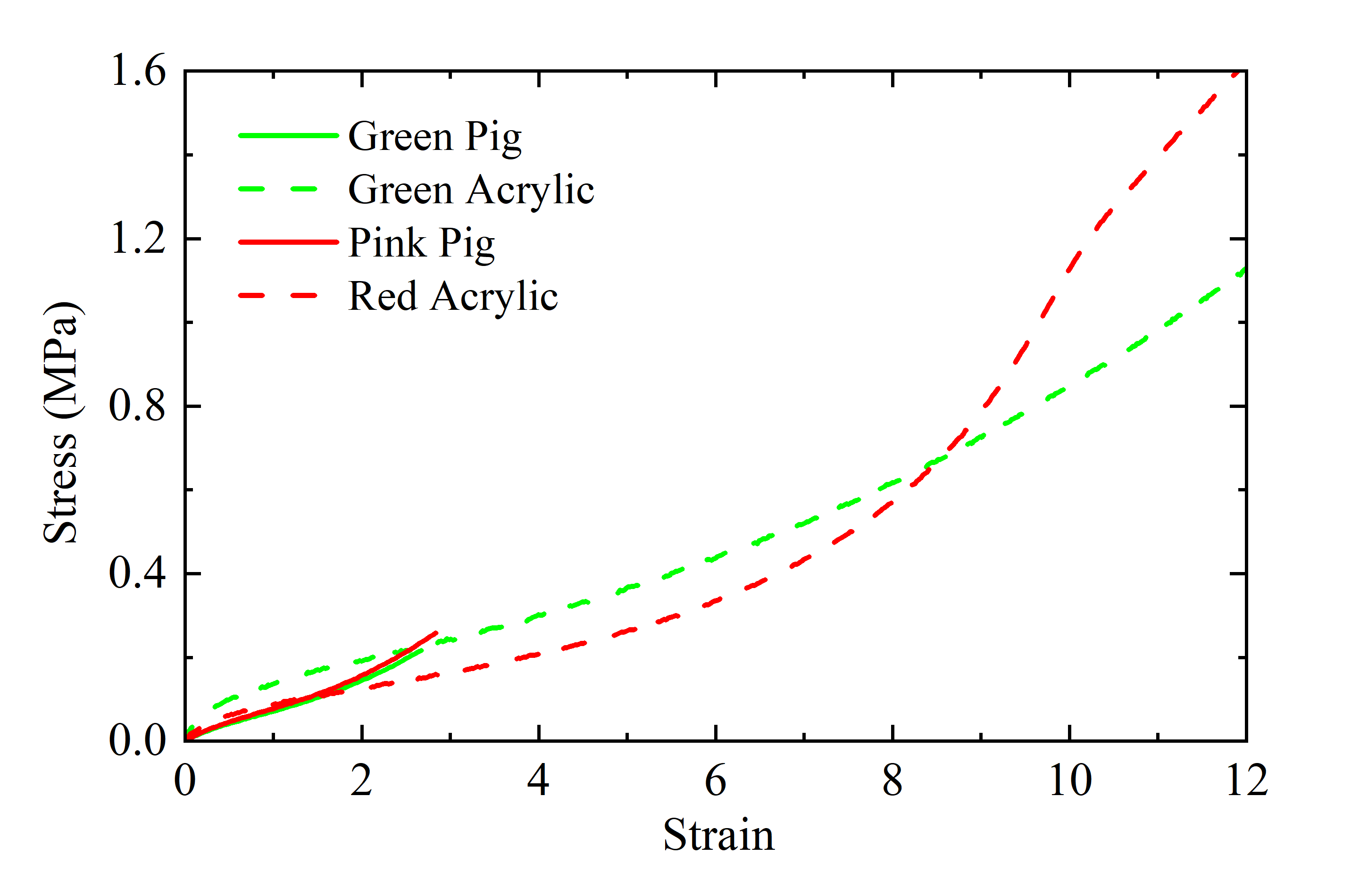}
    \vspace{-20pt}
	\caption{Stress-strain curves of the fluorescent silicone paint, where the acrylic paints represent the synthesized silicone adhesive based paints. The graphs have been cut off at the point of breakage of the dogbone tensile testing pieces.}
	\label{fig:stressssss}
	\vspace{-5pt}
\end{figure}

As shown in Fig. \ref{fig:stressssss}, we see that both of the synthesized (acrylic) paints outperformed the Silc Pig based fluorescent paints in percent elongation at breakage and their ultimate tensile strengths (UTS). Whereas the green and red acrylic paints gave us a percent elongation of 1,348\% and 1,212\%, and a UTS of 1.37 MPa and 1.65 MPa, respectively, the corresponding values for the Silc Pig paints were 267\%, 269\%, 0.23 MPa, and 0.26 MPa. 

We believe that these differences in values between the two types of paint could in part be due to the extensive amount of Silc Pig pigment we put into the silicone mixture. Although the pigment is synthesized to work well with and be miscible with silicone, the pigment has the potential to disrupt crosslink formation of the silicone, which would cause the silicone material to have lower percent elongation and UTS values. 

This phenomena could have also occurred with the addition of acrylic paint to the silicone adhesive. However, we believe that because silicone adhesive is designed to adhere to many different types of surfaces, disruption of crosslink formation would not have had as severe of an effect on the integrity of the synthesized paint. 

Another interesting effect we noticed was that the acrylic paints had different strain-stress curve behaviors. Although both curves behaved in similar ways at the beginning of the tensile test, the red acrylic paint began to exhibit a nonlinear behavior before breakage. We believe this behavior can be attributed to the material differences of the red and green acrylic paints from Liquitex and how miscible they are with the NOVOCS Gloss and silicone adhesive mixture. The Silc Pig mixtures, on the other hand, behaved similarly most likely because these pigments were manufactured to perform in this way.

With regards to illumination schemes, we found that although the acrylic paint on an acrylic sheet visually performed equivalently with the other two at discerning finer details, it was not better than the other two paints at 2D localization. In fact, the silicone adhesive acrylic paint mixture slightly outperformed all of the paints, but not significantly so. The reconstruction algorithm gave a 1.4 pixel error for the circle centers and a 85.0\% Dice score for the rectangular reconstruction segmentation image, compared to 3.6 pixel errors for both the Silc Pig and acrylic on acrylic and a 75.0\% and 72.3\% similarity coefficient, respectively. A comprehensive list of our results is shown in Table \ref{tab:bleh}. 

Overall, our metrics show that there is not a significant difference between the different types of illumination schemes. This metric could be different if we were to include the reconstruction normal gradients; however, we found that including normal gradients was not useful for many applications and that 2D localization was an acceptable alternative. 

We also note that for the acrylic on acrylic illumination scheme, there was more of a uniform lighting scheme throughout the entire sensor. Specifically, the tip of the sensor for both the Silc Pig paint and the synthesized paint had a slightly dimmer illumination at the tip compared to the base of the sensor. The nonuniform illumination, which did not affect our experiment results, is most likely due to the lower index of refraction of silicone compared to acrylic. As a result, more of the blue light will have a higher chance of refracting out of the silicone material instead of experiencing total internal reflection within the material, as it did with the blue LEDs pressed against an acrylic piece. 

\section{EXPERIMENT}
\subsection{Nut Classification}
To leverage the new, improved hardware modifications, we chose to utilize the Baby Fin Ray in an object ``sorting" task, using nuts that were still in their shells. This task shows how the new sensor can classify objects based on their shape and texture.  

\begin{figure}[ht]
	\centering
	\includegraphics[width=1.0 \linewidth]{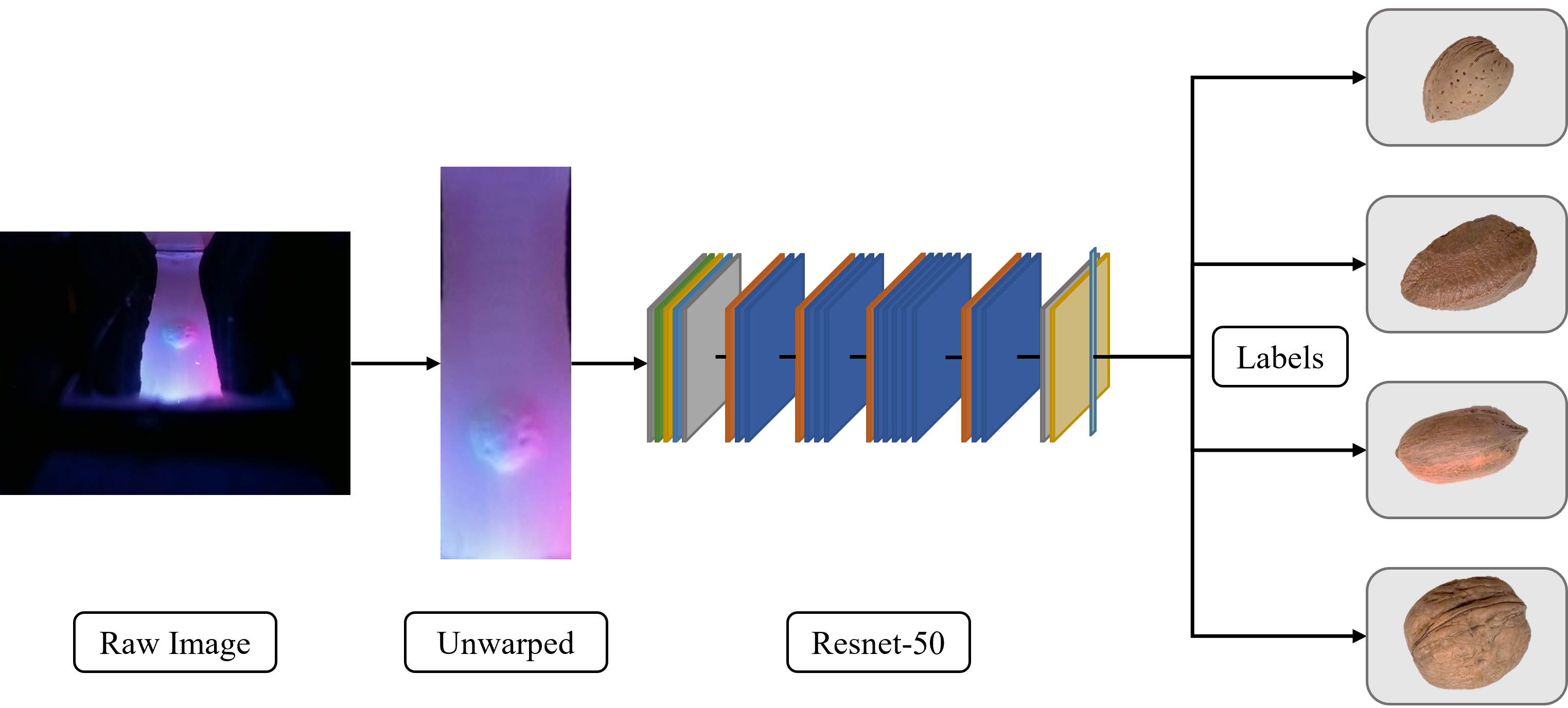}
    \vspace{-15pt}
	\caption{The architecture used for training our nut classification task. We take the raw image from our tactile sensor, unwarp the mirror image, before feeding it into a Resnet-50 architecture with four classes: Almond, Brazil Nut, Pecan, and Walnut.}
	\vspace{-5pt}
	\label{fig:resnet}
\end{figure}

For our task, we chose to use almonds, Brazil nuts, pecans, and walnuts, which all have semi-distinct shell textures and are of various shapes and sizes. We collected 500 tactile images each of the four different nuts, pressed against different parts of the sensor in various orientations and with flipped lighting configurations. The grasps were guided by hand. Next, the images were processed via an image unwarping process and the dataset was split into 80\% for training and 20\% for validation, with data augmentation applied on the training dataset. We used a Resnet-50 architecture, shown in Fig. \ref{fig:resnet}, for training on the unwarped images with four total classes representing the different types of nuts \cite{resnet}. For our neural net, we used the stochastic gradient descent optimizer with a learning rate of 1e-3, and a learning rate scheduler with a step size of 7 and a gamma of 0.1. 

We also attempted to train the same neural network on raw images instead of on our unprocessed and warped images. 

\subsection{Results}
Without unwarping, our neural network was able to achieve 99.5\% accuracy, while with unwarping, our neural net achieved a 95.8\% accuracy on the validation set. We believe that this slight discrepancy is due to the amount of ``negative" space in our unprocessed images. Although it was not immediately visible to our eyes, there are potential discrepancies in the black covering of the Fin Ray when different sized objects are being grabbed. As such, the ``negative" space might lead to higher accuracy for classifying the nut shells over this dataset, but could potentially have a lower accuracy on sensors that might be manufactured slightly differently.

Overall, our unwarped neural net had 98.1\%, 96.8\%, 96.9\%, and 91.3\% accuracy on classifying almonds, Brazil nuts, pecans, and walnuts, respectively. We also found that the Baby Fin Ray was able to both conform well to these differently sized nuts and was robust enough to withstand multiple digging attempts through our bowl of nuts. Not only did the synthesized paint stay bonded to the silicone, the design changes also allowed us to have more compliance and gave us more repeatable lighting results in between multiple grasps. The tapered end of our structure also means that the Baby Fin Ray could be used for digging tasks. These changes overall make the Baby Fin Ray more usable in manipulation tasks than its predecessor.

\section{CONCLUSION AND DISCUSSION}
The ability to leverage both the compliance and flexibility of soft robots with the intricate sensing afforded by camera-based tactile sensing is essential. If we hope to continue advancing soft manipulation so that soft grippers can eventually perform complex human manipulation tasks, we almost certainly need both of these abilities in our grippers. In this paper, we show an updated version of the GelSight Fin Ray, which not only encompasses both abilities, but has been improved so that it can potentially perform robustly in manipulation tasks. 

We increased its robustness by making hardware changes to the GelSight Fin Ray, including an update to the camera placement and a revamp of the illumination design for better compliance. Additionally, we specifically designed the finger for digging through clutter tasks. Our analysis of the different fluorescent paint illumination schemes also shows us that although there is not too much difference between the different schemes, our synthesized silicone adhesive and acrylic paint mixture showed the highest elongation at breakage and UTS. These properties show promise for the synthesis of other types of soft grippers with proprioception or tactile sensing, since there are many soft grippers that either expand or are constantly experiencing some form of deformation. 

Finally, we showcase that these new design implementations and sensing capabilities allow us to repeatedly grab nuts with the conformability of the Baby Fin Ray, and then use the camera-based sensor to very accurately classify between the various distinct textures. These abilities opens up avenues for us to, in the future, perform complicated digging tasks through cluttered household environments to retrieve specific objects. We also keep in mind that adding markers would allow us to perform incipient slip analysis and also allow us to potentially perform proprioception.

Overall, we show strategies to integrate more intricate, highly-detailed sensing into soft, compliant robots, and to do so without jeopardizing the original flexible and conformable capabilities of the soft robot. Doing so allows soft robotics and soft manipulation to progress even further so that they can one day perform at the level of human hands.

\section{ACKNOWLEDGEMENTS}

Toyota Research Institute, the Office of Naval Research, and the SINTEF BIFROST (RCN313870) project provided funds to support this work. The authors would also like to thank Xuanhe Li, Jerry Zhang, and Jialiang (Alan) Zhao for their advice and help with tensile testing, task design, neural net training, and general tips. Finally, the authors would like to show their appreciation for the insightful comments Arpit Agarwal and Yichen Li gave about lighting and reconstruction. 

\addtolength{\textheight}{-0cm}   



\bibliographystyle{IEEEtran}
\bibliography{Ref}
\end{document}